\documentclass{article}
\usepackage{ijcai19}

\usepackage{times}
\usepackage{soul}
\usepackage{url}
\usepackage[utf8]{inputenc}
\usepackage[small]{caption}
\usepackage{graphicx}
\usepackage{amsmath}
\usepackage{booktabs}
\usepackage{todonotes}
\usepackage{multirow}

\usepackage{amsthm}
\newtheorem{example}{Example}
\newcommand{\idest}{{\it i.e.}}

\newcommand{\etc}{{\it etc.}}
\newcommand{\etal}{{\it et al.}}
\newcommand{\bld}[1]{\textbf{#1}}

\newcommand{\dmodal}{\emph{deontic-modality}}
\newcommand{\dstruc}{\emph{deontic-structure}}
\newcommand{\dobj}{\emph{deontic-object}}
\newcommand{\objcond}{\emph{object-conditional}}
\newcommand{\Dmodal}{\emph{Deontic-modality}}
\newcommand{\Dstruc}{\emph{Deontic-structure}}
\newcommand{\Dobj}{\emph{Deontic-object}}
\newcommand{\Objcond}{\emph{Object-conditional}}

\newcommand{\emboff}[1][]{E_{\mathit{off#1}}} % Offset embedding
\newcommand{\embconc}[1][]{E_{\mathit{conc#1}}} % Concatenation embedding
\title{Classifying Norm Conflicts using Learned Semantic Representations}

\author{
João Paulo Aires$^1$\and
Roger Granada$^1$\and
Juarez Monteiro$^1$\and
Rodrigo C. Barros$^2$\and
Felipe Meneguzzi$^2$\\
\affiliations
Pontifical Catholic University of Rio Grande do Sul\\
\emails
$^1$ \{joao.aires.001, roger.granada, juarez.monteiro\}@acad.pucrs.br,
$^2$ \{rodrigo.barros, felipe.meneguzzi\}@pucrs.br,
}

\begin{document}

\maketitle

\begin{abstract} % REDO
While most social norms are informal, they are often formalized by companies in contracts to regulate trades of goods and services. 
When poorly written, contracts may contain normative conflicts resulting from opposing deontic meanings or contradict specifications. 
As contracts tend to be long and contain many norms, manually identifying such conflicts requires human-effort, which is time-consuming and error-prone. 
Automating such task benefits contract makers increasing productivity and making conflict identification more reliable. 
To address this problem, we introduce an approach to detect and classify norm conflicts in contracts by converting them into latent representations that preserve both syntactic and semantic information and training a model to classify norm conflicts in four conflict types. 
Our results reach the new state of the art when compared to a previous approach.

\end{abstract}

%%%%%%%%%%%%%%%%%%%%%%%%%%%%%%%%%%%%%%%%%%%%%%%%%%%%%%%%%%%%%%%%%%%%%%%%%%%%%%%%%%%%%%%%%%%%%%%
%%%%%%%%%%%%%%%%%%%%%%%%%%%%%%%         Introduction         %%%%%%%%%%%%%%%%%%%%%%%%%%%%%%%%%%
%%%%%%%%%%%%%%%%%%%%%%%%%%%%%%%%%%%%%%%%%%%%%%%%%%%%%%%%%%%%%%%%%%%%%%%%%%%%%%%%%%%%%%%%%%%%%%%
\section{Introduction}
\label{sec:introduction}
% Context.
Most societies use contracts as a central tool to formalize agreements~\cite{RousseauParks1993}, often dealing with the exchange of a service or goods offered by a creditor and paid by a debtor~\cite{HartHolmstrom1987}. 
Contracts are semi-structured documents that describe the agreement subject, its parties, and a series of definitions of what is expected from each party during the agreement validation. 
We can think about these definitions in terms of social norms~\cite{LuckEtAl2013} and formalize clauses that indicate the involved parties, a deontic meaning (obligation, prohibition, or permission), and an action to be performed (the object of the norm). 
Since contracts tend to be long and complex to cover as many situations that could arise out of an agreement\footnote{Contracts in our training corpus have an average of 6118 words.}, such complexity invariably leads to the danger of logical contradictions between the norms described in natural language, which in turn leads to norm conflicts. 
Norm conflicts are often the result of a clash between specifications, for example, a mistake in one norm might cause another norm to be impossible to comply with~\cite{ElhagEtAl2000}, and may invalidate a contract. 

% Motivation.
As natural language uses a diverse and often vague way to express ideas, identifying a norm conflict and its causes in contracts is a challenging task. 
An ever larger number of contracts being currently generated necessitates a fast and reliable process to identify norm conflicts. 
However, since such contracts are written in natural language, traditional revision methods involve contract makers reading the contract and identifying conflicting points between norms. 
Such a method requires huge human-effort and may not guarantee a revision that eliminates all conflicts. 
In response, we provide three contributions towards automatically identifying and classifying potential conflicts between norms in contracts. 
First, we formalize four types of norm conflicts in Section~\ref{sec:conf_types} based on existing work~\cite{Sadat2003} on the differences in deontic meaning and norm structure of contractual clauses. 
Importantly, our conflict typology is amenable to the learning tasks we develop in this paper.  
Second, we extend an existing corpus of norm conflicts~\cite{AiresEtAl2017dataset} with contracts and additional conflicting clauses. 
We annotate each conflict pair in the new dataset with its conflict type and describe this process in Section~\ref{sec:dataset}. 
Such addition and annotation allows us to identify complex conflicting cases involving small differences on norm structures and conditional terms. 
Third, in Section~\ref{sec:norm_classification}, we develop an approach based on sentence embeddings to classify norm conflicts according to our conflict typology. 
Unlike prior approaches to learning of norm conflicts, our contribution requires no parameter tuning to identify conflicts. 
We evaluate the resulting approach empirically in Section~\ref{sec:experiments}, showing that our results surpass the current state-of-the-art approach for classifying conflicts in contracts. 

%%%%%%%%%%%%%%%%%%%%%%%%%%%%%%%%%%%%%%%%%%%%%%%%%%%%%%%%%%%%%%%%%%%%%%%%%%%%%%%%%%%%%%%%%%%%%%%
%%%%%%%%%%%%%%%%%%%%%%%%%%%%%%%     Norms and Contracts      %%%%%%%%%%%%%%%%%%%%%%%%%%%%%%%%%%
%%%%%%%%%%%%%%%%%%%%%%%%%%%%%%%%%%%%%%%%%%%%%%%%%%%%%%%%%%%%%%%%%%%%%%%%%%%%%%%%%%%%%%%%%%%%%%%

\section{Norms and Contracts}
\label{sec:norms_contracts}

Societies often use norms to define behaviors commonly agreed by their members~\cite{AndersenTaylor2007}, with a wide range of norm applications from common sense to Government law~\cite{MahmoudEtAl2014}. 
Norms codify an expected behavior, which should be enforced, by defining in terms of deontic logic how and when certain actions should be performed~\cite{MahmoudEtAl2014}. 
Deontic logic has its origins on modal logic and deals with the notions of ``ideal'' worlds from the point of view of compliance with a body of stipulation worlds~\cite{CarmoJones2002}. 
These stipulations are the object of deontic modalities of prohibitions, obligations, and permissions. 

As contracts comprise series of norm statements specifying what each party is expected to fulfill, it is important that these statements are logically consistent. 
Any mistake in specifying the statements of the norms in a contract may lead to conflicts between them. 
This is particularly true for contracts in natural language since language may be ambiguous and writers of such contracts may overlook subtle logical conflicts. 
Therefore, understanding how these conflicts arise and what are their configurations is important for writing enforceable contracts. 
Norm conflicts are the result of a collision between two or more norms due to their stipulations of what ought to be done~\cite{NikiforakisEtAl2012}. 
As norms describe what is expected of the parties of a contract, they use deontic meanings (permission, prohibition, and obligation) to state how parties must behave in each situation. 
When a contract contains two norms whose simultaneous compliance is impossible, this contract contains a norm conflict. 
In such case, norms are mutually exclusive since compliance with one implies in noncompliance with the other, and thus, they cannot exist in a legal order~\cite{Sadat2003}. 

\citeauthor{Vranes2006}~\shortcite{Vranes2006} argues that a conflict arises in three main ways: (1) when exists an obligation to perform a certain action and a permission not to perform it; (2) when we have a prohibition to perform a given action and the permission to perform it; and (3) when involves an obligation and a prohibition.
\citeauthor{Sadat2003}~\shortcite{Sadat2003} expands \citeauthor{Vranes2006}' definitions by describing four causes for a norm conflict to arise. 
The first cause is when the same act is subject to different types of norms. 
Thus, a conflict between norms arises ``if two different types of norms regulate the same act, \idest, if the same act is both obligatory and prohibited, permitted and prohibited, or permitted and obligatory''. 
Example~\ref{ex:first_cause} illustrates a norm conflict between an obligation and a prohibition. 
\begin{example}
    \label{ex:first_cause}
    -
    \begin{enumerate}
        \item The receiving State shall exempt diplomatic agents from indirect taxes. 
        \item The receiving State shall not exempt diplomatic agents from indirect taxes. 
    \end{enumerate}
\end{example}
The second cause occurs when one norm requires an act, while another norm requires or permits a `contrary' act. 
In such case, a norm conflict occurs if ``two contrary acts, or if one norm permits an act while the other norm requires a contrary act''~\cite{Sadat2003}. 
Example~\ref{ex:sec_cause} illustrates the conflict, where both norms indicate different places in which a prisoner of war must be treated. 
Norm 1 states that it must be done in the prisoner camps, whereas norm 2 states that it must happen in civilian hospitals. 
The conflict arises when one tries to comply with one norm and is not complying with the other. 
\begin{example}
    \label{ex:sec_cause}
    -
    \begin{enumerate}
        \item Prisoners of war suffering from disease may be treated in their camps. 
        \item Prisoners of war suffering from disease shall be treated in civilian hospitals. 
    \end{enumerate}
\end{example}
The third cause for a conflict between norms is when one norm prohibits a `necessary precondition' of another norm. 
Suppose there are two actions A and B and action A has to be performed before action B. 
A norm conflict arises when one norm prohibits A and another norm allows B, as Example~\ref{ex:third_cause} illustrates. 
In the example, we consider action A as ``enter area X'' and action B as ``render assistance to any person in danger in area X''. 
\begin{example}
    \label{ex:third_cause}
    - 
    \begin{enumerate}
        \item Ships flying the flag of State A shall/may render assistance to any person in danger in area X. 
        \item Ships flying the flag of State A shall not enter area X. 
    \end{enumerate}
\end{example}
The fourth cause of norm conflict arises when one norm prohibits a `necessary consequence' of another norm. 
Suppose that one cannot perform action B without producing C as result. 
The conflict arises when one norm obliges B and another norm prohibits C, as Example~\ref{ex:fourth_cause} illustrates. 
If we consider action B as ``replace existing rails in area X'' and C as the period of time in which the line in area X will be hampered, one cannot comply with both norms 1 and 2 in Example~\ref{ex:fourth_cause}. 
\begin{example}
    \label{ex:fourth_cause}
    - 
    \begin{enumerate}
        \item State A shall replace existing rails with new ones in area X. 
        \item State A shall not hamper the transport of goods on the existing line in area X. 
    \end{enumerate}
\end{example}

%%%%%%%%%%%%%%%%%%%%%%%%%%%%%%%%%%%%%%%%%%%%%%%%%%%%%%%%%%%%%%%%%%%%%%%%%%%%%%%%%%%%%%%%%%%%%%%
%%%%%%%%%%%%%%%%%%%%%%%%%%%       Norm Conflict Types      %%%%%%%%%%%%%%%%%%%%%%%%%%%
%%%%%%%%%%%%%%%%%%%%%%%%%%%%%%%%%%%%%%%%%%%%%%%%%%%%%%%%%%%%%%%%%%%%%%%%%%%%%%%%%%%%%%%%%%%%%%%
\section{Norm Conflict Types}
\label{sec:conf_types}

% Rules for each name:
% deontic-modality: dmodal
% deontic-structure: dstruc
% deontic-object: dobj
% object-conditional: objcond

Automating the identification of potential conflicts between norms is the key process to make contract writing and revision faster. 
By identifying such cases in contracts, the contract writer can prioritize detailed revision of those norms identified as being most likely to be in conflict. 
In order to further help contract authors, we want to indicate what causes a conflict to arise within a contract, instead of simply detecting conflicting clauses in a contract.  
While there are various typologies for norm conflicts~\cite{Alchourron1991,VasconcelosEtAl2009,NikiforakisEtAl2012,RauhutWinter2017}, in the case of conflicts in contracts, one specific typology stands out by relating the deontic representation of norms within clauses to the possible types of conflicts~\cite{Vranes2006}. 
However, in order to diagnose the specific nature of the conflict and amend clauses to eliminate them, contract writers often need more information than the deontic modalities of two clauses that are in conflict. 
Thus, we leverage the typology of \citeauthor{Sadat2003}~\shortcite{Sadat2003} to identify four conflict types that can facilitate the task of eliminating existing conflicts by finely defining the \textit{nature} of the conflicts. 
The four types are: \dmodal, \dstruc, \dobj, and \objcond. 
Since we based our dataset on that by \citeauthor{AiresEtAl2017dataset}~\shortcite{AiresEtAl2017dataset}, we had to annotate all its norm conflicts. 
Its conflicts fit either the \dmodal\ or \dstruc~conflict types, thus, we were able to add them to our final dataset.

In order to formalize the four conflict types, consider a contract $\mathcal{C}$ and a set of norms $\mathcal{N} = \{n_{1}, n_{2}, ..., n_{n}\}$ where $\mathcal{N} \subseteq \mathcal{C}$ and $n_i$ is the $i^{th}$ norm in the contract.
Consider the four main elements of a norm as follows: Party, Deontic Meaning, Action, and Condition.
In our formalization, $\mathcal{P} = \{p_{1}, p_{2}, ..., p_{n}\}$ stands for the set of parties in $\mathcal{C}$ and $\mathcal{D} = \{d_1, d_2, d_3\}$ stands for the set of deontic meanings consisting of Permission, Prohibition, and Obligation.
$\mathcal{A} = \{a_1, a_2, ..., a_n\}$ stands for the set of actions declared in $\mathcal{C}$, and $Cond = \{c_1, c_2, ..., c_n\}$ is a set of conditions described in norms from $\mathcal{N}$. 
We add a second index to each element to describe the party, deontic meaning, action, and condition as part of a norm. 
For example, given a norm $n_i$, we represent the action $a_q$ expressed in such norm as $a_{q,i}$, where $q$ is the action identifier and $i$ is the norm identifier. 
Therefore, we can classify each conflict type. 

\Dmodal~conflict type represents the simplest conflict case.
It arises when two norms ($n_i$ and $n_j$) refer to the same parties ($p_{k,i}$ and $p_{k,j}$), express the same action ($a_{q,i}$ and $a_{q,j}$), but with different deontic meanings ($d_{r,i} \neq d_{t,j}$).
In norms written in English, the deontic meaning is often represented by a modal verb, such as \emph{must}, \emph{shall}, \emph{ought}, \etc~
Thus, \dmodal~conflict type consists of norms where modal verbs indicate different deontic meanings, subsuming most conflict types identified by~\citeauthor{Vranes2006}~\shortcite{Vranes2006}.
The following pair of norms illustrates how this conflict type can occur.
Here, \bld{may} indicates a permission whereas \bld{shall not} indicates a prohibition.
\begin{enumerate}
    \item The Specifications \bld{may} be amended by the NCR design release process.
    \item The Specifications \bld{shall not} be amended by the NCR design release process
\end{enumerate}

\Dstruc~conflict type is similar to \dmodal~conflict since it involves two norms ($n_i$ and $n_j$) with different deontic meanings ($d_{r,i} \neq d_{t,j}$) but with a different natural language structure. 
In this case, it can refer to the same subject (\idest, ($a_{q,i}$ and $a_{q,j}$), ($p_{k,i}$ and $p_{k,j}$), and ($c_{u,i}$ and $c_{u,j}$)) using different expressions. 
The following example illustrates this conflict type. 
\begin{enumerate}
\item All inquiries that Seller receives on a worldwide basis relative to Buyer's air chamber "Products" as specified in Exhibit III, \bld{shall} be directed to Buyer.
\item Seller \bld{may not} redirect inquiries concerning Buyer's air chamber "Products".
\end{enumerate}

This example contains two norms with different sentence structures involving the use of different word and word-structure referring to the same subject.
The conflict arises because in the first norm there is an obligation towards the Seller that inquiries must be sent to Buyer, whereas the second norm prohibits the Seller to send such inquiries.

\Dobj~conflict type indicates cases where both norms ($n_i$ and $n_j$) have the same deontic meaning ($d_{r,i}$ and $d_{r,j}$) but different overall natural language structures.
In this particular case, the conflict arises due to the definitions about the norm actions ($a_{q,i}$ and $a_{w, j}$) or specification details ($c_{u,i}$ and $c_{o,j}$).
The example below illustrates the case.
\begin{enumerate}
\item Autotote shall make available to Sisal one (1) working prototype of the Terminal by \bld{May 1}, 1998.
\item Autotote shall make available to Sisal one (1) working prototype of the Terminal by \bld{June 12}, 1998.
\end{enumerate}
In this case, although the action itself is the same in both norms, the conflict arises because their date definition differs in such a way as to allow the fulfillment of norm 2 while violating norm 1.
% A different case for this same conflict case is exemplified below.
The example below exemplifies a different (and more direct) instance of the same type of conflict. 
Specifically, we have a conflict because the actions are mutually exclusive.
\begin{enumerate}
    \item CoPacker will assume no costs of transportation and handling for such rejected Products.
    \item CoPacker shall assume all costs of transportation and handling rejected Products.
\end{enumerate}

\Objcond~conflict type between a pair of norms ($n_i$ and $n_j$) occurs when a condition or exception in one norm ($c_{u,i}$) conflicts with the action expressed in the second norm ($a_{q,j}$).
In this case, the deontic meanings, parties, and actions can be either the same or different. 
This type of conflict concerns specific examples where a condition is involved, and the following example illustrates the case.

\begin{enumerate}
\item The Facility shall meet all legal and administrative code standards applicable to the conduct of the Principal Activity thereat.
\item Only if previously agreed, the Facility ought to follow legal and administrative code standards.
\end{enumerate}

In this example, we have two norms that have the same deontic meaning, but the condition in the second one creates a conflict since one can comply to it and, if it was not previously agreed, do not follow legal and administrative code standards for facilities.

%%%%%%%%%%%%%%%%%%%%%%%%%%%%%%%%%%%%%%%%%%%%%%%%%%%%%%%%%%%%%%%%%%%%%%%%%%%%%%%%%%%%%%%%%%%%%%%
%%%%%%%%%%%%%%%%%%%%%%%%%%%       Dataset construction. %%%%%%%%%%%%%%%%%%%%%%%%%%%
%%%%%%%%%%%%%%%%%%%%%%%%%%%%%%%%%%%%%%%%%%%%%%%%%%%%%%%%%%%%%%%%%%%%%%%%%%%%%%%%%%%%%%%%%%%%%%%

\section{Dataset Extension}
\label{sec:dataset}

Given the difficulty in finding any sizable corpus with actual norm conflicts that is openly available, we use the existing norm conflict dataset organized by \citeauthor{AiresEtAl2017dataset}~\shortcite{AiresEtAl2017dataset} as basis for our dataset. 
Since their dataset does not contain any annotation for the types of conflicts, we performed its annotation by identifying one of the four types of conflicts (\dmodal, \dstruc, \dobj, and \objcond) for each pair of norms marked as conflicting. 
By annotating the dataset we observed that it includes only two types of conflicts (\dmodal\ and \dstruc). 
Thus, we extend the existing datasets with further conflicts. 
We use an approach similar to the semi-automated described by \citeauthor{AiresMeneguzzi2017coin}~\shortcite{AiresMeneguzzi2017coin}. 
Specifically, we developed a web-based tool (Available at http://lsa.pucrs.br/concon/).
that randomly selects norms within the contract corpus and creates a copy of them. 
A human volunteer, who understands the types of norm conflicts, is instructed to freely edit the copied norm in order to create a conflict with the original norm, following one of the four types described earlier in the paper. 
By deliberately inserting conflicts into the contract we ensure that the new contract has a conflicting pair of norms in it allowing the creativity of the volunteers to add variability to the specific language that leads to a conflict. 
We did not evaluated or post-processed created conflicts since we assume volunteers correctly generated them. 
The resulting dataset contains a total of 228 conflicting norms including the existing 111 conflicts from the previous dataset in addition to a total of 11,329 non-conflicting sentence pairs. 

\begin{table}[tb]
    \centering
    %\scriptsize
    \begin{tabular}{l|c|c}
        \toprule
        \bld{Conflict Type} & \bld{\# of Elements} & \bld{\% in dataset}\\
        \midrule
        \dmodal     & 97    & 42\%\\
        \dstruc     & 61    & 27\%\\
        \dobj       & 30    & 13\%\\
        \objcond    & 40    & 18\%\\
        \bottomrule
        % Total     & 228 & 100\%\\
        % \hline
    \end{tabular}
    \caption{Number and proportion of each conflict type in the new dataset}
    \label{tab:dataset}
\end{table}

Table~\ref{tab:dataset} details the number of conflicts of each type in our new dataset, where we can see that the proportion of \dmodal~conflict type is the highest one. 
This occurs because it is the simplest type of conflict one can create. 
On the other hand, creating a conditional case is rather difficult since one needs to find a plausible condition that fits the norm context. 

%%%%%%%%%%%%%%%%%%%%%%%%%%%%%%%%%%%%%%%%%%%%%%%%%%%%%%%%%%%%%%%%%%%%%%%%%%%%%%%%%%%%%%%%%%%%%%%
%%%%%%%%%%%%%%%%%%%%%%%%%%%       Norm Conflict Classification      %%%%%%%%%%%%%%%%%%%%%%%%%%%
%%%%%%%%%%%%%%%%%%%%%%%%%%%%%%%%%%%%%%%%%%%%%%%%%%%%%%%%%%%%%%%%%%%%%%%%%%%%%%%%%%%%%%%%%%%%%%%
\section{Classification of Norm Pairs}
\label{sec:norm_classification}

Identification of conflicts is a binary classification problem, where a pair of norms is classified either as a conflict or non-conflict, which previous work has addressed with various degrees of success~\cite{AiresEtAl2017norm,AiresMeneguzzi2017coin,AiresEtAl2018}. 
Nevertheless, previous research never attempts to identify the nature of the conflicts it identifies/ 
We address this limitation by expanding the binary classification to a multi-class classification task, where we classify a pair of norms according to a predefined conflict typology or as non-conflict. 
Unlike conflict identification, in conflict classification we want not only to identify whether a norm pair is a conflict but also to classify its type. 
Each pair of norms containing a conflict can be classified into one of the four types of conflicts we propose in Section~\ref{sec:conf_types}: \dmodal, \dstruc, \dobj, or \objcond.

Before classifying norms pairs, we transform each norm written in natural language within a contract into a vector representation that captures its semantic information. 
According to \citeauthor{LeMikolov2014}~\shortcite{LeMikolov2014}, word embeddings can capture linguistic regularities, \idest, the semantic similarity of words is captured by the similarity of the corresponding vector representation. 
Recent research has extended the notion of word embedding to capture semantic information from sequences of words rather than individual words using an efficient unsupervised learning algorithm to train their distributed representations called Sent2Vec~\cite{PagliardiniEtAl2018}. 
Specifically, Equation~\ref{eqn:sent2vec} computes the sentence embedding $v_S$ for the sentence $S$, where $R(S)$ is the list of words in the sentence, by averaging the embeddings of its constituent words. 
Finally, in order to improve generality, Sent2Vec performs random sub-sampling, \idest, deleting random words once Sent2Vec extracts all the unigrams.

%\vspace{-1mm}
\begin{equation}
    v_S = \frac{1}{|R(S)|} \sum\limits_{w \in R(S)} v_w
    \label{eqn:sent2vec}
\end{equation}
%\vspace{-1mm}

Once we train a Sent2Vec encoder for the contract dataset, we convert each norm in a contract into its corresponding sentence embedding before classification. 
In order to classify pairs of norms, we use a supervised approach based on either the concatenation $\embconc$ or the offset $\emboff$ of embeddings representing pairs of norms. 
\citeauthor{AiresEtAl2018}~\shortcite{AiresEtAl2018} shows that offset embeddings can achieve high performance in conflict identification between two sentences (norms) by learning a vector representing this conflict as the following vector offset:
    
%\vspace{-1mm}
\begin{equation}
    v_{\text{conflict}} = \frac{1}{|\mathcal{P}|} \sum\limits_{(n_1, n_2) \in \mathcal{P}} v_{n_1} - v_{n_2}
    \label{eq:conflictOffset}
\end{equation}

\noindent
where $\mathcal{P}$ represents the set of all norm pairs that contain conflicts, and $v_{n_1}$ and $v_{n_2}$ are the vector embeddings of each norm of the pair. 
In this work, instead of using all conflicts to generate a single embedding vector, we use the offset of each norm pair. 

In order to train our classifier, we extract pairs of norms ($n_1$, $n_2$) from our dataset, where $n_1$ conflicts with $n_2$. 
We compute the embedding ($v_{n_1}$, $v_{n_2}$) of each norm pair using the Sent2Vec library\footnote{\url{https://github.com/epfml/sent2vec}} and a pre-trained model from the Wikipedia English corpus, and use these embeddings to compute the offset vector ($v_{\text{offset}}$). 
The number of pairs representing non-conflicting norms in our dataset (Section~\ref{sec:dataset}) is much larger than the number of pairs representing types of conflicting norms. 
Due to the unbalanced nature of our dataset, we use two types of classifier. 
The first classifier uses 5 classes, describing a non-conflict and four conflict types. 
The second classifier uses 4 classes describing only the conflicting pairs (\dmodal, \dstruc, \dobj, or \objcond).  
By narrowing classification only to conflicting pairs, we can shed light on what types are the hardest and the easiest to classify by training an SVM classifier using either the concatenation  $\embconc$ or the offset $\emboff$ of embeddings representing norm pairs. 

%%%%%%%%%%%%%%%%%%%%%%%%%%%%%%%%%%%%%%%%%%%%%%%%%%%%%%%%%%%%%%%%%%%%%%%%%%%%%%%%%%%%%%%%%%%%%%%
%%%%%%%%%%%%%%%%%%%%%                    Experiments                  %%%%%%%%%%%%%%%%%%%%%%%%%
%%%%%%%%%%%%%%%%%%%%%%%%%%%%%%%%%%%%%%%%%%%%%%%%%%%%%%%%%%%%%%%%%%%%%%%%%%%%%%%%%%%%%%%%%%%%%%%
\section{Experiments}
\label{sec:experiments}

In order to perform experiments, we divided our dataset into $k$-folds and test set, where each fold contains an equal number of pairs of norms representing conflicts and non-conflicts.  
We train and validate our models using $k$-fold cross validation ($k=10$), and select the model trained in the fold with the highest F-score to use against the test set. 
We measure the performance of each model in terms of Accuracy ($\mathcal{A}$), Precision ($\mathcal{P}$), Recall ($\mathcal{R}$) and F-measure ($\mathcal{F}$). 

As classifier, we use an off-the-shelf implementation of a Support Vector Machine with the implementation of \citeauthor{CrammerSinger2001}~\shortcite{CrammerSinger2001} for multiclass classification from \emph{scikit-learn}\footnote{http://scikit-learn.org} toolbox. 
We do not optimize any parameter, using the default values from the toolbox (L2 penalty, Square hinge loss, C=1.0). 
The training step consists of learning an equal number of embeddings representing conflicts and non-conflicts. 
The test step predicts classes for a norm embedding of unseen pairs of norms. 

\subsection{Results}
\label{sec:quantitative}

\begin{figure}[t!]
    \centering
    \includegraphics[width=.45\textwidth]{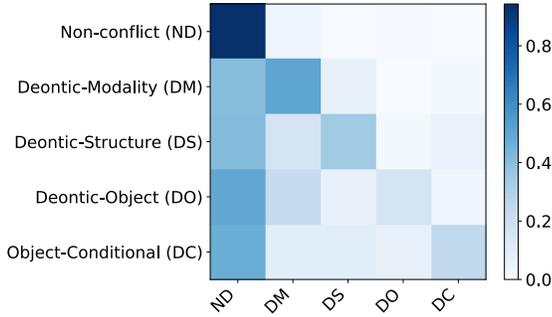}
    \caption{Confusion matrix of the \emph{TypeC+Non} ($\embconc$) model.}
    \label{fig:cm_all}
\end{figure}

Due to space limitations, we report only the scores achieved using the test set\footnote{Code, experiments, and results available at https://bit.ly/2E7MSUB.}, as shown in Table~\ref{tab:our_approaches}, where \emph{TypeC+Non} identifies the classification using 5 classes (4 types of conflicts + non-conflicts) and \emph{TypeC} identifies the classification of norms using only the four types of conflicts, $\embconc$ is the classification using the concatenation of embeddings, and  $\emboff$ the classification using their offset. 

\begin{table}[h]
    \centering
    \small
    \caption{Performance summary, where `TypeC+Non' is the classification using conflicts and non-conflicts and `TypeC' is the classification using only conflicts.}
    \begin{tabular}{l|c|c|c|c}
        \toprule
            \bld{Approach} & $\mathcal{A}$ & $\mathcal{P}$ & $\mathcal{R}$ & $\mathcal{F}$ \\
        \midrule
            TypeC+Non ($\emboff$)         &      0.63  &      0.62 &       0.38  &      0.42  \\
            TypeC+Non ($\embconc$)        &      0.70  &      0.71 &       0.64  &      0.66  \\
            TypeC ($\emboff$)             &      0.55  &      0.43 &       0.40  &      0.40  \\
            TypeC ($\embconc$)            & \bld{0.77} & \bld{0.80} & \bld{0.75} & \bld{0.75} \\
            \cite{AiresMeneguzzi2017coin} &      0.63  &      0.59  &      0.64  &      0.61  \\
        \bottomrule
    \end{tabular}
    \label{tab:our_approaches}
\end{table}

\begin{figure}[t!]
    \centering
    \includegraphics[width=.45\textwidth]{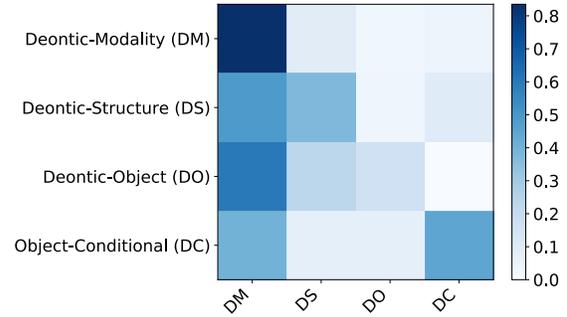}
    \caption{Confusion matrix of the \emph{TypeC} ($\embconc$) model.}
    \label{fig:cm_multi_conf}
\end{figure}

As Table~\ref{tab:our_approaches} exhibits, we achieve 77\% accuracy when classifying only norm pairs that have conflicts (\emph{TypeC} model), decreasing to 70\% when including non-conflicts in the target classes. 
Since our dataset is unbalanced, \idest, pairs of norms are not equally distributed into classes, these results provide evidence that the classifier may be biased towards the non-conflict class, which contains half of the pairs of norms. 
Analyzing the distribution of correct classified pairs of \emph{TypeC+Non} ($\embconc$) using a confusion matrix illustrated in Figure~\ref{fig:cm_all}, we note that most of the corrected classified pairs are non-conflicting norms. 
When generating the confusion matrix for the \emph{TypeC} ($\embconc$) model as illustrated in Figure~\ref{fig:cm_multi_conf}, we can see that this model tends to classify most conflicts as \dmodal. 
Indeed, Table~\ref{tab:dataset} corroborates the intuition that this bias is likely due to more than half of the pairs containing conflicts (52\%) being of the \dmodal~class. 

In order to compare our approach to the current state of the art, we train existing approaches with our new dataset. 
We compare using the approach of \citeauthor{AiresMeneguzzi2017coin}~\shortcite{AiresMeneguzzi2017coin}, since the approach that uses offset~\cite{AiresEtAl2018} can only be used for conflict identification. 
Using the code from \citeauthor{AiresMeneguzzi2017coin} to perform identification using a CNN, we modified their last layer of the network to perform a multiclass classification. 
Table~\ref{tab:our_approaches} compares the results of their approach to ours and shows that our approach obtains better results (by a substantial margin) than the state of the art approach that uses a CNN~\cite{AiresMeneguzzi2017coin}. 

\subsection{Qualitative Analysis}
\label{sec:qualitative}

We evaluate the effectiveness of the approaches qualitatively by selecting specific examples of correct and incorrect classifications that illustrate the nature of our \emph{TypeC+Non} classifier against the previous state of the art~\cite{AiresMeneguzzi2017coin}.
Since in our test set, \citeauthor{AiresMeneguzzi2017coin}'s approach only correctly classified examples from Non-conflict and Deontic-Modality types, we limit our comparison to these two types for this approach. 

\citeauthor{AiresMeneguzzi2017coin}'s approach correctly classifies the following norm pair as belonging to the \dmodal~type with 98\% of confidence. 
Our approach reports 50\% of confidence to the same type, showing that our classifier may be impacted by the length of the sentence. Thus, in long sentences, our approach decrease the confidence of classifying conflicts correctly. 

\begin{enumerate}
    \item The terms of this Letter Agreement shall become effective immediately prior to the closing under the APA.
    \item The terms of this Letter Agreement must not become effective immediately prior to the closing under the APA.
\end{enumerate}

As previously described, this type of conflict has the modal verb as its main difference.
This high similarity between norms facilitates the classification for \citeauthor{AiresMeneguzzi2017coin}'s approach since it performs its classification over the character similarity of norms.

Finally, we selected the following pair of norms that were misclassified as non-conflicting by \citeauthor{AiresMeneguzzi2017coin}'s approach with 64\% confidence, and correctly classified by our approach with 40\% confidence. 

\begin{enumerate}
    \item Invoice cost shall not be adjusted for, and Customer shall not be entitled to, promotional allowances, cash discounts, prompt pay discounts, growth programs or any other supplier incentives received by USF.
    \item If the cost of the invoice is adjusted, Customer will not be entitled to promotional discounts, cash discounts, growth programs or any other supplier incentives received by USF.
\end{enumerate}

Although this is a simple example of \objcond~conflict type, \citeauthor{AiresMeneguzzi2017coin}'s approach considered it a non-conflict. 
One of the reasons can be the difference in the beginning of both norms.
We notice a pattern of classification in this approach that tends to classify as conflict norms that begin with the same sequence of characters. 
On the other hand, our approach could identify the meaning of \objcond~conflict type with a certain margin to the second highest probability 27\% \dobj~conflict type. 

The probabilities for each conflict type from our test set suggests that our approach tends to spread the probabilities across conflict types. 
For example, the highest probability assigned to a conflict type was 50\% with the other 50\% spread on the other conflict types. 
Although our approach achieves better results to conflict type classification, it does not provide high confidence in the classification as the \citeauthor{AiresMeneguzzi2017coin}'s approach provides. 

%%%%%%%%%%%%%%%%%%%%%%%%%%%%%%%%%%%%%%%%%%%%%%%%%%%%%%%%%%%%%%%%%%%%%%%%%%%%%%%%%%%%%%%%%%%%%%%
%%%%%%%%%%%%%%%%%%%%%%%%%%%%%%%%      Related Work      %%%%%%%%%%%%%%%%%%%%%%%%%%%%%%%%%%%%%%%
%%%%%%%%%%%%%%%%%%%%%%%%%%%%%%%%%%%%%%%%%%%%%%%%%%%%%%%%%%%%%%%%%%%%%%%%%%%%%%%%%%%%%%%%%%%%%%%
\section{Related Work}
\label{sec:related_work}

Recent research focuses on automating contract processing. 
\citeauthor{ChalkidisAndroutsopoulos2017}~\shortcite{ChalkidisAndroutsopoulos2017} developed two approaches to extract and classify contract elements (termination dates, legislation references, contracting parties, and agreed payments). 
The first approach automatically extracts elements using bidirectional Long-Short Term Memory (BiLSTM) and classifies them using an LSTM neural network. 
The second approach combines a BiLSTM neural network with a Conditional Random Field (CRF) classifier to classify contract elements. 
Their experiments use a dataset of approximately 3,500 English contracts, annotated with 11 types of contract elements, and achieve an overall F-score of 87\% using the modified BiLSTM.
\citeauthor{ChalkidisEtAl2018}~\shortcite{ChalkidisEtAl2018} developed a second contract processing approach to extract obligations and prohibitions from contracts. 
They use word embedding to train a hierarchical BiLSTM and classify norm elements using 6 different classes consisting of different norm structures for obligations and prohibitions.

We can divide recent approaches on the specific task of norm conflict identification into two broad classes. 
The first class of approaches involves a, largely manual, translation of the contract from natural language into a controlled language amenable to automated processing. 
The second class works directly on natural language to identify conflicts. 
Among the approaches using a controlled language, \citeauthor{RossoEtAl2011}~\shortcite{RossoEtAl2011} develop a formal contract language to convert natural language norms and detect conflict using rules.
To test their approach, they create a conflicting example and use their rules to identify the conflicts. 
\citeauthor{AzzopardiEtAl2016}~\shortcite{AzzopardiEtAl2016} develop a similar approach, which translates natural language norms into a deontic logic representation and identify conflicting cases with a deontic logic reasoner. 
They identify some conflict cases, such as permission and prohibition, obligation and prohibition, obligation of two mutually exclusive actions, and obligation and permission of mutually exclusive actions. 
These conflicts largely fit our notion of \textit{deontic-modality} and \textit{deontic-object} conflicts. 
The main difference between this type of approach and ours is that we try to obtain all information needed to identify conflicts directly through natural language, and we are capable of identifying a richer set of conflict types. 
Thus, instead of converting norms to a controlled language, we use an embedding approach that retains the semantic information. 

Using information directly from natural language, \citeauthor{AiresMeneguzzi2017coin}~\shortcite{AiresMeneguzzi2017coin} propose a convolutional neural network (CNN) to classify norm pairs as conflict or non-conflict. 
They convert norm pairs into a binary matrix by setting the characters of one norm as rows and the characters of the other as columns, then they set 1's to cells where row and column have the same character.
They use Aires~\etal~dataset to test their approach and obtain 84\% of accuracy.
Finally, the closest approach to ours is described by \citeauthor{AiresEtAl2018}~\shortcite{AiresEtAl2018}, which detects norm conflicts using an offset vector generated by the embedding of conflict sentences. 
In order to identify a conflict, they compare the distance generated by the subtraction of a pair of norm embeddings with an offset vector containing the average value of differences between all conflict norms. 
If the distance is below a threshold ($\lambda$), the pair of norms is considered a conflict.  
Our approaches have two distinct advantages over theirs.
First, we can identify not only that there are conflicts between norms, but also the type of conflict, providing additional help for contract reviewers. 
Second, and most importantly, we do not have to manually select a threshold ($\lambda$) to identify the conflict, instead, we use embeddings as input to an SVM that identifies and classifies norm conflicts.

%%%%%%%%%%%%%%%%%%%%%%%%%%%%%%%%%%%%%%%%%%%%%%%%%%%%%%%%%%%%%%%%%%%%%%%%%%%%%%%%%%%%%%%%%%%%%%%
%%%%%%%%%%%%%%%%%%%%%%%%%%%%%%%%%      Conclusion      %%%%%%%%%%%%%%%%%%%%%%%%%%%%%%%%%%%%%%%%
%%%%%%%%%%%%%%%%%%%%%%%%%%%%%%%%%%%%%%%%%%%%%%%%%%%%%%%%%%%%%%%%%%%%%%%%%%%%%%%%%%%%%%%%%%%%%%%
\section{Conclusion}
\label{sec:conclusion}

In this paper, we developed an approach to classify conflicts between norms in contracts. 
Our approach consists of manipulations on embedding representations of norms in order to identify conflicts. 
As part of our contribution, we defined four conflict types to classify a conflict and help on conflict solving and extended an existing norm conflict corpus adding the new types and used it to train norm-conflict classifiers. 
Compared to existing approaches, our approach based on embeddings surpasses the state of the art approach adapted for classification tasks. 
Such result shows that using comparative embeddings has a powerful impact on identifying conflicts. 

As future work, we have 4 main goals. 
First, we aim to gather more data to increase the the frequency of each conflict type relative to each other. 
Second, we want to indicate directly on the text what parts of the norm are causing the conflict, enabling us to highlight the specific language that lead to the conflict. 
Third, using the identification of conflict types, we aim to perform automatic conflict resolution. 
Finally, we aim to evaluate our approach with human contract writers in order to quantify how valuable such tool really is for them. 

\section*{Acknowledgement}
The authors would like to thank CAPES/FAPERGS for partially funding this research. We gratefully acknowledge the support of NVIDIA Corporation with the donation of the Titan Xp GPU used for this research.

% \section*{Acknowledgments}
% Omitted due to blind review.
% \bibliographystyle{named}
% \bibliography{ijcai19}

\end{document}